# An Integrated Optimization and Deep Learning Pipeline for Predicting Live Birth Success in IVF Using Feature Optimization and Transformer-Based Models

Arezoo Borji[1,2,3], Hossam Haick[4,5], Birgit Pohn[6], Antonia Graf[6], Jana Zakall[6], S M Ragib Shahriar Islam[2,1,3], Gernot Kronreif[2], Daniel Kovatchki[7], Heinz Strohmer[7], Sepideh Hatamikia[1,2*]

1 Research Center for Clinical AI-Research in Omics and Medical Data Science (CAROM), Department of Medicine, Danube Private University (DPU), Krems, Austria
2 Austrian Center for Medical Innovation and Technology, Wiener Neustadt, Austria
3 Department of Medical Physics and Biomedical Engineering, Medical University of Vienna, Vienna, Austria
4 Laboratory for Life Sciences and Technology (LiST), Department of Medicine, Danube Private University, Krems, Austria
5 Department of Chemical Engineering, Technion – Israel Institute of Technology, Haifa, Israel
6 Department of Medicine, Danube Private University (DPU), Krems, Austria
7 Kinderwunschzentrum an der Wien, Austria

**Abstract**

In vitro fertilization (IVF) is a widely utilized assisted reproductive technology, yet predicting its success remains challenging due to the multifaceted interplay of clinical, demographic, and procedural factors. This study develops a robust artificial intelligence (AI) pipeline aimed at predicting live birth outcomes in IVF treatments. The pipeline uses anonymized data from 2010 to 2018, obtained from the Human Fertilization and Embryology Authority (HFEA). We evaluated the prediction performance of live birth success as a binary outcome (success/failure) by integrating different feature selection methods, such as principal component analysis (PCA) and particle swarm optimization (PSO), with different traditional machine learning-based classifiers including random forest (RF) and decision tree, as well as deep learning-based classifiers including custom transformer-based model and a tab transformer model with an attention mechanism. Our research demonstrated that the best performance was achieved by combining PSO for feature selection with the TabTransformer-based deep learning model, yielding an accuracy of 99.50% and an AUC of 99.96%, highlighting its significant performance to predict live births. This study establishes a highly accurate AI pipeline for predicting live birth outcomes in IVF, demonstrating its potential to enhance personalized fertility treatments.

## 1. Introduction

Assisted reproductive technologies (ART), particularly in vitro fertilization (IVF), have transformed the landscape of infertility treatment, offering hope to millions of couples worldwide [1]. Despite advancements in embryology and clinical practices, achieving consistent success in IVF remains a significant challenge [2]. Key outcomes, such as success in embryo implantation and live birth, depend on a multitude of factors; the complexity of IVF outcomes therefore stems from the intricate interplay of numerous factors that must align for a successful treatment [3, 4]. This includes patient age, hormonal profiles, clinical protocols, embryological characteristics, and even lifestyle or genetic factors, all of which contribute to the multifaceted nature of the process [5]. Each of these variables influences treatment success, making it challenging to predict outcomes and optimize protocols. As illustrated in Figure1, the IVF process involves key stages, each contributing to these outcomes, from patient evaluation and ovarian stimulation to embryo selection and transfer. Traditional methods for embryo selection and live birth prediction are often unable to integrate and analyze these multidimensional data aspects effectively as they primarily rely on static morphological grading systems, while foundational, are often subjective and limited in their ability to capture the complex dynamics of embryonic development and live birth outcomes [3].

Recent advancements in machine learning and artificial intelligence (AI) have introduced a paradigm shift in IVF, providing tools to analyze vast and complex datasets with unprecedented precision [4]. AI and ML have revolutionized IVF by automating embryo evaluation, predicting implantation potential, and enhancing live birth outcomes [5]. These technologies address many of the limitations of traditional methods, offering unprecedented precision, consistency, and scalability. They enable the analysis of large and complex datasets, offering predictive insights that surpass the capabilities of traditional statistical models [6].

One of the most promising applications of AI in IVF is embryo selection, where AI can predict the likelihood of a live birth for individual embryos. Deep learning models, especially convolutional neural networks (CNNs), have shown remarkable success in automating embryo grading by analyzing time-lapse imaging data [8]. This technology helps identify embryos with a higher probability of resulting in a live birth, significantly enhancing decision-making during the IVF process. Annotation-free scoring systems, such as those described by Ueno et al. [7], have further streamlined the embryo evaluation process by eliminating the need for extensive manual input while maintaining high predictive accuracy. These models analyze morphogenetic parameters, such as pronuclear fading, cleavage patterns, and blastulation timing, providing dynamic insights into embryo development that were previously unattainable through static morphological assessments [8].

Beyond embryo grading, AI has been employed to predict implantation potential with notable success. Machine learning algorithms, such as random forests and ensemble models, integrate morphokinetic data and patient characteristics to assess the likelihood of implantation. Studies by Bamford et al. [9] and Uyar et al. [10] have demonstrated the ability of these models to achieve area under the curve (AUC) values exceeding 0.75 for implantation prediction. Furthermore, reinforcement learning-based systems like Dyn Score could dynamically update predictions in real time, offering clinicians actionable insights into embryo viability [11]. These adaptive models represent a significant step forward in IVF decision-making, allowing for more personalized and precise treatment strategies.

The goal of IVF is to achieve a live birth, making the prediction of live birth outcomes a critical focus of AI research [2]. AI models, which integrate patient demographics, clinical data, and simple quantitative features from imaging modalities, have shown promise in this domain. For instance, studies by Huang et al. [12] and Jiang et al. [13] utilized voting ensembles of CNNs to predict embryo ploidy, resulting in significant improvements in live birth rates. These models enable clinicians to optimize treatment protocols and maximize the likelihood of success.

Feature selection techniques also have been proposed to develop efficient AI-based methods to support the IVF process. Kragh et al.[14] explores distinctions between ranking embryos based on implantation potential and predicting probabilities of implantation success, as well as issues like dataset balancing, selection bias, and clinical applicability. By focusing on the most relevant features, these methods enhance model interpretability and reduce computational complexity without compromising performance. Studies by Ueno et al. [7] and Bamford et al. [9] have highlighted the importance of feature selection in improving the efficiency and accuracy of IVF-based predictive models.

Several studies proposed promising AI methods for classifying live birth success as a binary outcome (success/failure). For example, Zhang et al. [15] employed an artificial neural network (ANN) model, McLernon et al.[16] applied a discrete-time logistic regression model, while Jones et al. [17] also utilized logistic regression. Sanders et al. [18] conducted a comparison of live birth rates using binary logistic regression. Raful Hassan et al. [19] used a hill climbing wrapper algorithm for feature selection. Milewski et al. [20] employed the SIMBAF algorithm, a margin-based feature selection method that enhances classification performance. Finally, Different approaches achieved an accuracy between 0.73 to 0.96.

Despite the promising results achieved so far, the prediction of live birth outcomes using AI has not yet been integrated into clinical practice, underscoring the need for further innovation and development of more robust approaches in this area. Most prior relevant research on live birth prediction has primarily relied on traditional AI models, often overlooking the significant performance enhancements that advanced deep learning methodologies could offer [15-20]. Leveraging these cutting-edge deep learning techniques has the potential to refine predictive accuracy and enable more reliable, data-driven decision-making in clinical settings.

Our work tries to enhance previous works on live birth prediction by presenting a novel, integrated optimization and deep learning pipeline designed to predict live birth success in IVF with greater accuracy. This pipeline seamlessly combines Particle Swarm Optimization (PSO), a metaheuristic optimization method, for feature selection with an advanced tab transformer-based deep learning model, offering an innovative and effective approach to tackling the complexity of IVF datasets. While PSO has been widely utilized in other fields for optimizing feature subsets, its potential in IVF prediction tasks remains largely untapped. By incorporating PSO, the pipeline identifies the most influential features, streamlining the model and enhancing its interpretability. Simultaneously, transformers, originally developed for natural language processing, are adapted to capture intricate interactions between clinical and demographic variables, demonstrating superior predictive capabilities compared to traditional machine learning models. The use of transformer models for IVF prediction tasks, including live birth prediction, remains an unexplored area of research. This study shows combination of PSO, and transformers provides a robust framework with significant performance for advancing IVF live birth prediction. We validated the proposed method using the open access dataset 2010-2018 HFEA.

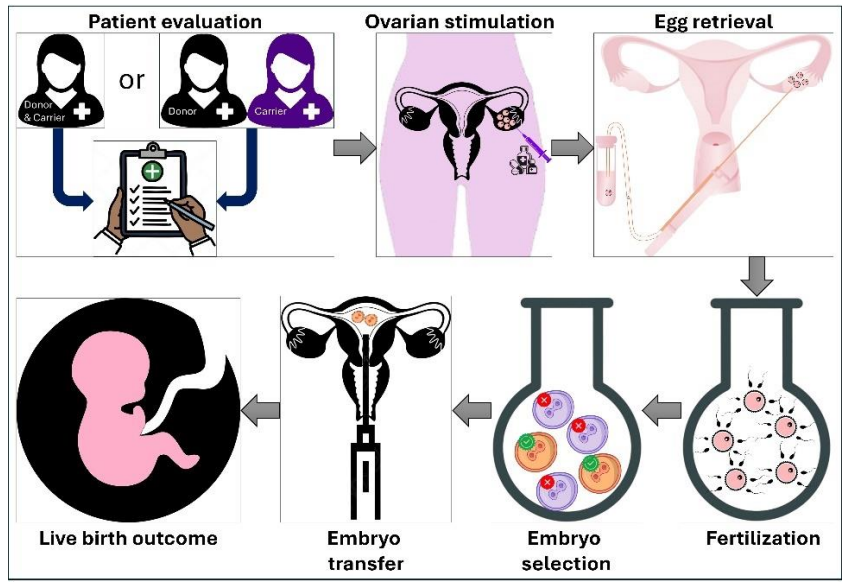

Figure 1. Step-by-step process of in vitro fertilization (IVF).

Figure 1 shows the IVF procedure, starting with patient assessment for appropriateness and donor egg utilization. After hormone-stimulated ovarian stimulation, ultrasound-guided aspiration retrieves several eggs. The eggs are fertilized in the lab, then embryo selection selects the healthiest embryos for transfer. Lastly, the uterus receives the selected embryos for successful implantation and a live birth. It shows the precision and complexity of assisted reproductive technology.

## 2. Methodology
### 2.1. General experimental design

In this work, we have applied inclusion and exclusion criteria to enhance the quality and relevance of the data, ensuring it was appropriate for our binary classification task. To reduce the dimensionality of the dataset and improve model performance, we utilized two feature selection and reduction techniques: Principal Component Analysis (PCA) and PSO. For classification, we evaluated the performance of four different classifiers: Random Forest (RF), Decision Tree (DT), a transformer-based model, and a tab transformer-based model. Finally, we designed different experimental setups: the first used PCA features as input to all classifiers (Method-1 and Method-3, Figure 2), and the second used features provided by PSO as input to all these classifiers (Method-2 and Method-4, Figure 2). In total, we have eight classification models including PCA+RF, PCA+Decision Tree, PSO+RF, PSO+Decision Tree, PCA+Transformer based model, PCA+ Tab-transformer based model, PSO+Transformer based model, PSO+Tab-transformer based model.

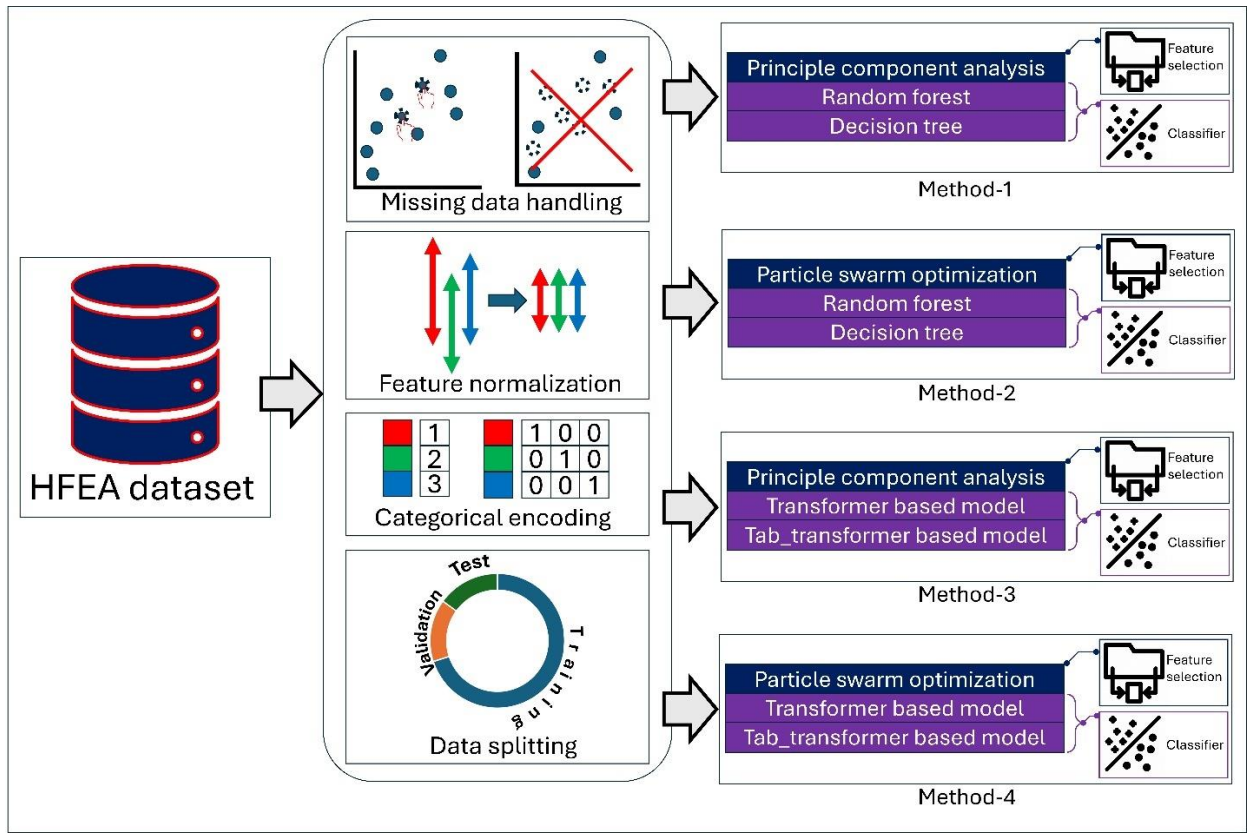

Figure 2. Overview of preprocessing steps and classification methods used in this paper

### 2.1.1. The dataset used

For this study, we utilized the Human Fertilization and Embryology Authority (HFEA) dataset, an anonymized registry dataset that encompasses fertility treatments conducted from 2010 to 2018. Designed to enhance patient care and maintain strict confidentiality for patients, donors, and offspring, this dataset stands as one of the most comprehensive and longest-running repositories of fertility treatment records globally. With 665,244 patient records and an initial set of 94 features, it provides a detailed account of fertility treatment cycles, covering patient demographics, treatment protocols, and infertility causes. These attributes present a comprehensive view of the factors influencing IVF outcomes during this period.

The dataset includes both numerical and categorical variables, capturing a broad spectrum of critical factors. Key features encompass patient-specific details such as age at the time of treatment, number of prior IVF pregnancies, live birth outcomes, and specific infertility causes (e.g., tubal disease, ovulatory disorders, or male infertility factors). Additionally, we meticulously have recorded procedural details such as the type of eggs and sperm used (e.g., fresh, frozen, donor, or patient-derived), the number of eggs collected, and the number of embryos transferred. This level of granularity allows for an in-depth analysis of the variables affecting IVF success rates. The prediction performance of live birth success as a binary outcome (success/failure) is assessed in this study.

### 2.2. Data preprocessing pipeline for IVF data analysis

The preprocessing pipeline transformed raw IVF data (Section 2.1.1) into a clean and structured format for the AI pipeline's input. It began with column standardization, ensuring uniformity by converting names to lowercase and removing whitespace. We structurally aligned the datasets by reindexing and adding missing columns, and then consolidated them into a single DataFrame for holistic analysis. We removed columns with less than 1% non-null values to enhance data quality. We imputed missing values based on the data type. Finally, we numerically encoded categorical features and normalized numerical features to a [0, 1] range (Figure 2).

### 2.3. Inclusion and exclusion criteria

Inclusion and exclusion criteria were defined to ensure a clean, complete, and relevant dataset for this study. These criteria were chosen based on the groundwork laid by Sadegh-Zadeh et al. [21], who used the same dataset and adhered to a set of inclusion and exclusion criteria conditions for data preparation and analysis. The dataset included subjects who met the following conditions: (1) they had valid (non-missing) values for the target variable, "Live birth occurrence"; (2) they provided data for at least one infertility-related cause, such as "ovulatory disorder"; and (3) their cycle history indicated a non-negative record of prior treatment cycles. These criteria ensured the inclusion of relevant cases with sufficient data for analysis. We applied these inclusion criteria and then implemented exclusion criteria to improve the quality of the data. We excluded subjects with missing information for "elective single embryo transfer", as this variable was crucial for the analysis. Additionally, we removed entries with logical inconsistencies, such as negative treatment cycles or conflicting treatment-related dates, to ensure data validity. By adopting these inclusion and exclusion criteria, this study ensured a high-quality dataset suitable for robust predictive modeling of IVF live birth outcomes. The number of subjects included in this study after exclusion criteria is 115,012.

### 2.3.1. Feature selection using particle swarm optimization (PSO)

Feature selection reduces the number of predictors and focuses on the most relevant ones [22]. In this study we have employed PSO as a feature selection method due to its efficient search for optimal solutions in large and complex spaces [23]. PSO is a nature-inspired optimization technique, modeled after the social behavior of bird flocking or fish schooling. Each individual component, called a "particle," represents a candidate solution in the search space. These particles move through space by adjusting their positions based on their own experiences and those of neighboring particles, mimicking how animals in groups share information to find food or navigate environments [19]. PSO is particularly effective for solving complex optimization problems, including feature selection, where it can efficiently explore the vast combinatorial space of possible feature subsets. This study employs PSO to pinpoint the ideal feature subset for forecasting the success of live births. Each particle encoded a subset of features as a binary vector, where 1 indicated inclusion of a feature and 0 indicated exclusion of a feature.

**Cost Function**

The cost function in PSO evaluates the quality of each particle's solution (equation 1). In this study, the cost function is defined as below:

$$C = -(F1 - P \cdot N) \tag{1}$$

Where C is the cost function value to be minimized by PSO, F1 is the F1-score of a logistic regression model are trained on the selected features. The F1 score balances precision and recall, making it suitable for imbalanced datasets like IVF outcomes. P is the penalty weight, a parameter controlling the trade-off between model performance and simplicity. N is the number of features selected by the particle. The goal of PSO is to minimize C, which indirectly maximizes the F1 score while penalizing larger feature subsets. This ensures the final feature set is both performant and understandable. The following steps outline how to select features using PSO:

Algorithm 1: Pseudo code for the feature selection using PSO

---

**Inputs:**
 - Dataset with features: $F$
 - Cost function: $-(F1 - P \cdot N)$
 - Parameters: swarm size: $S = 20$, Maximum iterations: $T = 1000$, Inertia weight: $w = 0.7$, Cognitive acceleration coefficient: $c_1 = 1.5$, social acceleration coefficient: $c_2 = 2$, penalty factor: $P$.

**Outputs:**
 - Optimal feature subset: $F_{optimal}$
 - Best fitness value: $C_{best}$

**Procedure:**
1. **Initialization**:
   For each particle $i = 1, 2, ..., S$:
    - Initialize binary position vector $x_i \in \{0,1\}^n$ and velocity $v_i \in R^n$
    - Evaluate fitness: $C_i = -(F1 - P \cdot N)$, $N = \sum_j^n x_i |j|$
   Where $x_i |j|$ indicates whether features $j$ is selected.
    - Set the personal best position: $p_{best_i} = x_i$
   **End for**
   Set the global best position: $g_{best} = \arg \min_{i=1}^{S} C_i$

2. **Optimization**:
   While $t \leq T$:
    For each particle $i = 1, 2, ..., S$:
     For each feature $j \in \{1, 2, ..., n\}$:

- Update the velocity:
$$v_i|j| = w \cdot v_i|j| + c_1 \cdot r_1 \cdot (p_{best_i}|j| - x_i|j|) + c_2 \cdot r_2 \cdot (g_{best}|j| - x_i|j|)$$
Where $r_2, r_2 \sim$ Uniform (0,1) are random value.
- **Update position using a sigmoid function:**
  **If** sigmoid $(v_i|j|) = \frac{1}{1+e^{-v_i|j|}}$
**End for**
- Evaluate fitness: $C_i = -(F1 - P \cdot N), N = \sum_{j=1}^{n} x_i|j|$, update personal best:
**If** $C_i < C_{p\_best_i}$, then $p\_best_i = x_i$, $C_{p\_best_i} = C_i$
**End for**
Update the global best:

**If** $C_i < C_{g\_best_i}$, then $g_{best} = x_i$, $C_{g\_best} = C_i$
Increment the iteration counter: $t = t + 1$
**End while**
3. **Output**:
Return: $F_{optimal} = g_{best}, C_{best} = C_{g\_best}$

### 2.3.2. Dimensionality reduction with Principal Component Analysis (PCA)

Principal Component Analysis (PCA) is a statistical technique used to reduce the dimensionality of a dataset while retaining as much information as possible [24]. It does this by transforming the original data into a new set of orthogonal components, called principal components, which are ranked according to their ability to capture the variance within the data. In this study, we have applied PCA to the IVF dataset to reduce its dimensionality while retaining 95% of the data's variance. This process can remove irrelevant variations and reduce the computational complexity.

### 2.3.3. Random forest

Random Forest (RF) is an ensemble learning method that combines the outputs of multiple decision trees to improve predictive performance and reduce overfitting [25]. RF naturally evaluates feature importance by measuring the impact of each feature on prediction quality [26]. In this study, we have utilized a RF with 200 decision trees as estimators, each with a maximum depth of 10. Note that we restrict the depth of each tree to avoid overfitting and maintain interpretability. Moreover, criterion='gini' used Gini impurity to evaluate split quality.

### 2.3.4. Decision tree

A Decision Tree is a supervised learning algorithm used for classification and regression tasks [27]. It recursively splits the data into subsets based on feature thresholds, forming a tree-like structure where each internal node represents a decision based on a feature, and each leaf node represents an output prediction. Decision Trees are highly interpreted, as they clearly outline the decision-making process, making them particularly useful for understanding feature importance and validating selected features. In this study, we used the feature-extracted PCA and fed it into the decision tree with the following parameters: max_depth=10, limited depth to maintain simplicity, and criterion=gini.

### 2.3.5. Transformer-based model

A deep learning model based on the transformer architecture, known as a transformer-based model for classification, can solve classification tasks [28]. Vaswani et al. originally introduced transformer architecture in the "Attention Is All You Need" paper [29], and it has since become the foundation of many state-of-the-art models in natural language processing (NLP), computer vision, and other fields. The attention mechanism is a critical component of the transformer-based model, designed to analyze and interpret tabular data to predict IVF success. The attention mechanism enables the model to dynamically assign importance to specific features, capturing intricate relationships between them and improving the model's decision-making process. In this work, the dataset includes features such as patient age, sperm quality, number of embryos transferred, and other clinical parameters. These features often interact in complex ways. The attention mechanism dynamically determines which features are most important for predicting IVF success and adjusts their importance based on the context of the input data for each individual case. For instance, the attention mechanism may prioritize features such as the quality of embryos for older patients. Younger patients may receive more emphasis on features like the number of eggs retrieved.

The attention mechanism in this transformer-based model operates in the following steps:

**Step 1: Input transformation**

The input data is composed of tabular features, which are referred to as input_dim features after feature selection. We treat each feature as a component of the input vector. The features are first projected into a higher-dimensional space using a dense layer to make them suitable for attention computation:

$$X = \text{Dense}(x) \qquad (2)$$

After this, a sequence dimension is added to simulate sequential processing.

**Step 2: Scaled dot-product attention**

The scaled dot-product attention mechanism computes the relationships between features:

$$\text{Attention }(Q, K, V) = \text{softmax}\left(\frac{Qk^T}{\sqrt{d_k}}\right)v \qquad (3)$$

Where Query (Q) represents the feature being queried, key (K) represents the importance of each feature relative to the query, and value (V) contains the actual feature data.

Each feature attends to all other features, producing a matrix of attention scores that capture dependencies between them. The softmax function ensures that the attention scores sum to 1, creating a probabilistic weight for each feature.

**Step 3: Multi-head attention**

This work employs multi-head attention, dividing the input into multiple "heads." Each head learns to focus on different types of relationships. For instance, one individual might concentrate on the correlations between patient age and success. Another head might emphasize sperm quality or treatment type. We have concatenated and transformed the outputs from all heads into a single vector, combining multiple perspectives.

**Step 4: Residual connection and layer normalization**

The input is added back to the attention output:

$$x = \text{Add}(x, \text{AttentionOutput}) \qquad (4)$$

**Layer normalization:** The output is normalized to stabilize gradients and ensure smooth learning.

The attention mechanism powers the Transformer model, offering a sophisticated approach to tabular data analysis for IVF success prediction. The architecture of this transformer-based model proposed in this study is explained in the Table 1.

Table 1. The architecture of the proposed transformer-based classification model for predicting live birth success in IVF.

| Layer | Output Shape | Explanation |
| --- | --- | --- |
| Input Layer | (batch_size, 45) | Raw input features (45 selected features). |
| Dense Layer | (batch_size, 128) | Projects feature into 128 dimensions. |
| Sequence Expansion | (batch_size, 1, 128) | Adds a sequence dimension for Transformer processing. |
| Multi-Head Attention | (batch_size, 1, 128) | Learning relationships between features with 4 attention heads. |
| Residual + Normalization | (batch_size, 1, 128) | Preserves input information and normalizes activations. |
| Feed-Forward Network | (batch_size, 1, 128) | Further processes feature representations. |
| Residual + Normalization | (batch_size, 1, 128) | Adds stability and preserves the input. |
| Global Average Pooling | (batch_size, 128) | Aggregates the sequence into a single feature vector. |
| Dense (Output Layer) | (batch_size, 1) | Outputs a probability for the binary classification task. |

This work configures the transformer model with selected hyperparameters to optimize its performance on IVF success prediction. The number of selected features from PSO, representing the length of the reduced feature set (45 features), determines the input dimension (input_dim). We set the number of attention heads (num_heads) to 4, which enables the model to learn diverse relationships between features through parallel attention mechanisms. The feed-forward network dimension (ff_dim) is 128, providing a hidden layer size that refines feature representations after attention. The model includes two transformer encoder layers (num_layers), each consisting of a multi-head attention block and a feed-forward network enabling hierarchical feature extraction. The model applies to a dropout rate (dropout_rate) of 0.3 and L2 regularization (l2_reg) with a strength of $1e^{-4}$ to prevent overfitting. We set the batch size (batch_size) to 2048 during training to ensure efficient utilization of computational resources, and we set the learning rate (learning_rate) to a low value of $1e^{-6}$ to ensure stable and gradual convergence. The binary crossentropy loss function optimizes the model for binary classification tasks, monitoring performance metrics such as accuracy and AUC throughout the training process. The model limits training to a maximum of 40 epochs and implements an early stopping patience of 2 epochs to halt training if the validation loss does not improve, thereby preventing overfitting. These hyperparameters collectively ensure the model's ability to generalize well while capturing complex relationships within the IVF dataset. We have provided all these hyperparameter values using the grid search method. In Table 2, all these hyperparameters along with their respective values are shown.

Table2: The information of all parameters used for transformer model in this study.

| Parameter Name | Value |
| --- | --- |
| Input Dimension (input_dim) | 40 features |
| Number of Heads (num_heads) | 4 |
| Feed-Forward Dimension (ff_dim) | 128 |
| Number of Layers (num_layers) | 2 |
| Dropout Rate (dropout_rate) | 0.3 |
| L2 Regularization (l2_reg) | $1e^{-4}$ |
| Batch Size (batch_size) | 2048 |
| Learning Rate (learning_rate) | $1e^{-6}$ |
| Loss Function | Binary_crossentropy |
| Number of epochs | 40 |

### 2.3.6. Tab transformer-based model

The TabTransformer model is a deep learning approach that combines structured datasets with a mix of categorical and numerical features [30]. The tab transformer uses self-attention mechanisms to capture dependencies between features, particularly among categorical features. Instead of representing categorical data using traditional encoding methods, it maps each category to a learned embedding vector. These embeddings allow the model to capture semantic relationships between categories. The architecture starts by converting categorical features into embeddings and merging them with numerical features, either directly or via normalization layers. Transformer layers receive these inputs and use self-attention mechanisms to model the interactions between features. By doing so, the model identifies complex relationships between features that may be critical for the task, such as correlations between specific categories or numerical ranges. Table 3 summarizes the detailed architecture of the proposed tab transformer model in this study.

Table 3. The architecture of the proposed tab transformer-based classification model for predicting live birth success in IVF.

| Layer | Input Shape | Output Shape | Description |
| --- | --- | --- | --- |
| **Numerical Input Layer** | (batch_size, num_numerical) | (batch_size, num_numerical) | Input layer for scaled numerical features. |
| **Categorical Input Layer** | (batch_size, 1) (per feature) | (batch_size, embedding_dim) | Embedding layers convert categorical indices into dense vector representations. |
| **Concatenation Layer** | Combined inputs | (batch_size, total_dim) | Numerical features and categorical embeddings are concatenated. |
| **Reshape Layer** | (batch_size, total_dim) | (batch_size, 1, total_dim) | Reshapes the input for attention layers. |

| **Multi-Head Attention** | (batch_size, 1, total_dim) | (batch_size, 1, total_dim) | Self-attention layer captures feature dependencies and relationships. |
|---|---|---|---|
| **Layer Normalization** | (batch_size, 1, total_dim) | (batch_size, 1, total_dim) | Normalizes outputs from the attention mechanism for stable learning. |
| **Dense Layer (ReLU)** | (batch_size, 1, total_dim) | (batch_size, 1, 128) | Fully connected dense layer with ReLU activation to learn higher-level patterns. |
| **Output Layer** | (batch_size, 1, 128) | (batch_size, 1) | Sigmoid activation outputs probability for binary classification. |

Here, similar to transformer-based model proposed in Section 2.3.5., the attention mechanism plays a crucial role in learning complex relationships between the input features. It works by dynamically focusing on different parts of the feature set, which helps the model capture interactions between both numerical and categorical features. The multi-head attention mechanism is particularly powerful in this case because it allows the model to simultaneously attend to multiple aspects of the input data, learning different relationships in parallel.

The attention mechanism operates by computing attention scores that determine how much weight each feature should have in relation to others. Each feature is transformed into a query, key, and value, and the attention mechanism compares the query to all keys to compute a score. This score decides how much attention should be given to the corresponding value. By applying multi-head attention, the model can learn different types of relationships across features simultaneously. For instance, one head may focus on the interaction between numerical features, while another head could focus on categorical feature interactions. The explanation of all steps in the proposed tab transformer model are shown in Figure 3.

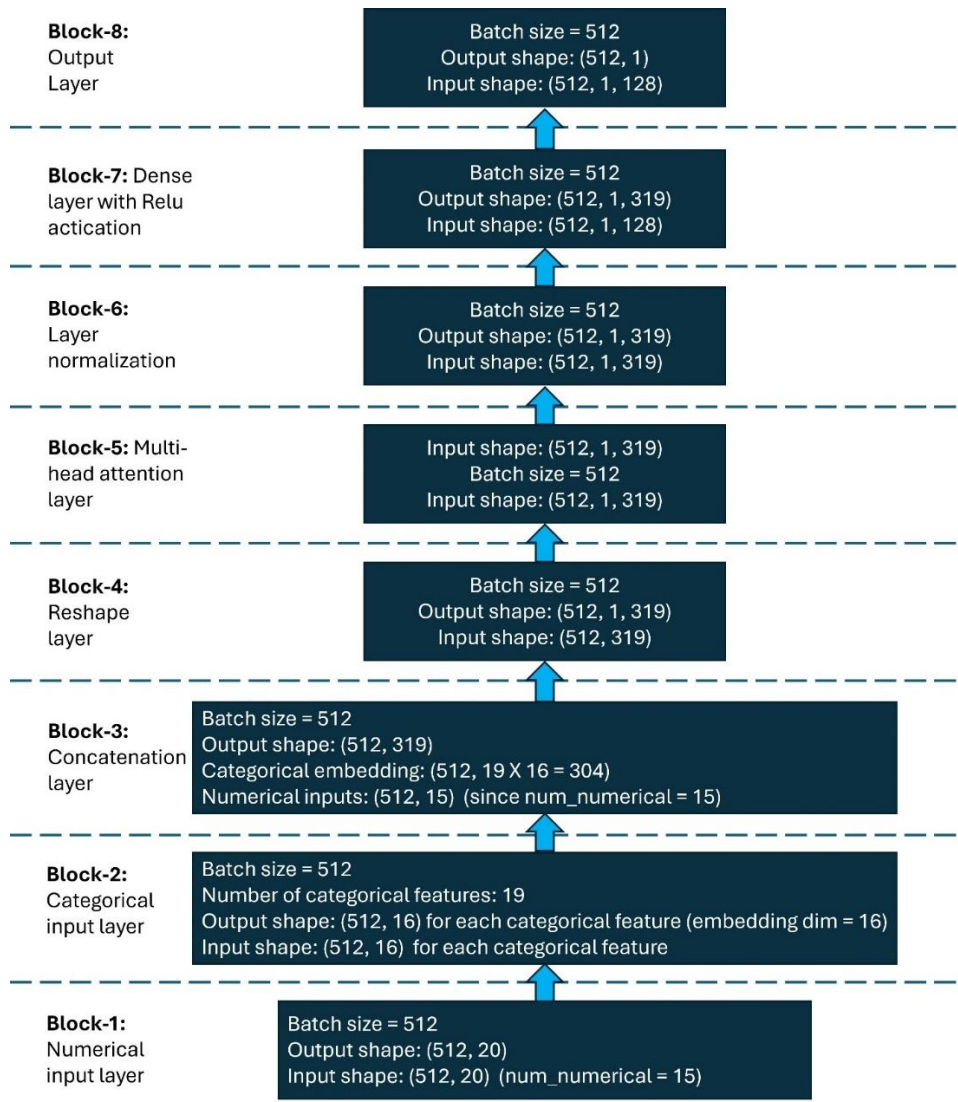

Figure 3. Overview of the proposed tab transformer model.

In the tab transformer architecture, various parameters define the model's structure and behavior. The model uses ReLU activation for the feedforward layers to introduce non-linearity. In the output layer, the activation function is sigmoid for binary classification tasks and softmax for multi-class classification tasks. During training, we employed the Adam optimizer with a learning rate of 0.0000001. The input shape for the numerical features is (batch_size, input_dim), where input_dim corresponds to the number of numerical features selected from the dataset. For categorical features, we encode each categorical feature as an integer index before passing it through the embedding layers, resulting in an input shape of (batch_size, 1) for each categorical column.

The multi-head attention mechanism has four attention heads. This allows the model to focus on multiple aspects of the data simultaneously, learning complex feature relationships. The feed-forward layers, which process the output of the attention mechanism, set the feed-forward dimension (ff_dim) to 128 units. The model consists of 1 layer of multi-head attention followed by feed-forward layers, with a dropout rate set to 0.2 to help prevent overfitting during training. To further reduce overfitting, we apply L2 regularization with a regularization strength of 0.01 to the weights in the feed-forward layers. To fine-tune the model, we set the batch size during training to 512 and the learning rate for the Adam optimizer to an extremely low value of 0.0000001. For binary classification tasks, we use binary cross-entropy as the loss function for training, while we typically use categorical cross-entropy for multi-class classification problems. The optimizer trains the model for 40 epochs, iteratively adjusting the weights to minimize the loss. For binary classification tasks, binary cross-entropy is used as the loss function. Table 4 summarizes the parameters and their corresponding values used for the proposed tab transformer model.

Table 4: The information of all key parameters used for the proposed tab transformer-based model and their respective values used in this study.

| Parameter | Value |
| --- | --- |
| **Activation Function (Feedforward)** | ReLU |
| **Activation Function (Output Layer)** | Sigmoid (binary), Softmax (multi-class) |
| **Optimizer** | Adam |
| **Learning Rate** | 0.0000001 |
| **Input Shape (Numerical Features)** | (batch_size, input_dim) |
| **Input Shape (Categorical Features)** | (batch_size, 1) |
| **Input Dimension (input_dim)** | Number of selected numerical features |
| **Number of Heads (num_heads)** | 4 |
| **Feed-Forward Dimension (ff_dim)** | 128 units |
| **Number of Layers (num_layers)** | 1 layer of multi-head attention + feed-forward layers |
| **Dropout Rate (dropout_rate)** | 0.2 |
| **L2 Regularization (l2_reg)** | 0.01 |
| **Batch Size (batch_size)** | 512 |
| **Loss Function** | Binary Cross-Entropy (binary classification), Categorical Cross-Entropy (multi-class classification) |
| **Number of Epochs** | 40 |

## 2.4. Performance evaluation metrics

Computational resources used in this study included an Intel(R) Core (TM) i7-10700K CPU running at 3.80 GHz, 32 GB of RAM, and an NVIDIA GeForce RTX 3080 GPU with 10 GB of VRAM. This hardware setup enabled efficient implementation of PSO and the transformer-based model, ensuring rapid experimentation and testing. We evaluated our machine learning models using several metrics, each providing distinct insights into model performance. These metrics include accuracy, precision, recall, and F1-score, which are defined as follows:

**Accuracy**: Accuracy measures the proportion of correctly predicted instances out of the total instances. It is calculated as:

$$Accuracy = \frac{TP+TN}{TP+FP+FN+TN} \quad (5)$$

Where:

TP: True Positives (correctly predicted positive cases)

TN: True Negatives (correctly predicted negative cases)

FP: False Positives (incorrectly predicted positive cases)

FN: False Negatives (incorrectly predicted negative cases)

**Precision**: Precision quantifies the proportion of correctly predicted positive cases out of all predicted positives. It is defined as:

$$Precision = \frac{TP}{TP+FP} \quad (6)$$

**Recall**: Recall, also known as sensitivity or true positive rate, measures the proportion of actual positives correctly identified by the model. It is calculated as:

$$\text{Recall} = \frac{TP}{TP+FN} \quad (7)$$

**F1-Score**: The F1-score is the means of precision and recall, offering a single metric that balances the two. It is calculated as:

$$F_1\_score = 2 \times \frac{precision \times recall}{precision+recall} \quad (8)$$

By employing these evaluation metrics, we obtained a comprehensive understanding of the models' strengths and weaknesses [31]. For all experiments, 10-fold cross validation was used in order to reduce overfitting and to ensure that the model generalizes well to unseen data.

# 3. Results
## 3.1. Classification results

Table 5 presents the validation results of eight classification models (Section 2.1) designed to predict live birth success in IVF. The performance of each model is reported by five performance metrics including accuracy, precision, recall, F1-score, and AUC.

Table 5. Performance evaluation of classification models for predicting live birth success in IVF. PCA: Principal Component Analysis, PSO: Particle Swarm Optimization, RF: Random Forest.

| The number of experiments | Classification Method | Accuracy | Precision | Recall | F1-Score | AUC |
|---|---|---|---|---|---|---|
| 1 | PCA + RF | 92.2% | 94% | 90% | 92.4% | 0.94 |
| 2 | PCA + Decision Tree | 91% | 89% | 88% | 95% | 0.90 |
| 3 | PSO + RF | 93.5% | 94.9% | 91.6% | 95.8% | 0.93 |
| 4 | PSO + Decision Tree | 91.9% | 90% | 91.6% | 94% | 0.93 |
| 5 | PCA + Transformer-Based Model | 95.5% | 96.7% | 93.4% | 96.5% | 0.95 |
| 6 | PCA + Tab_transformer-Based Model | 96.2% | 98.3% | 94.1% | 96.2% | 0.97 |
| 7 | PSO + Transformer-Based Model | 98.1% | 97.9% | 98.4% | 98.2% | 0.98 |
| 8 | PSO + Tab_transformer-Based Model | **99.50%** | **99.6%** | **99.5%** | **99.5%** | **0.99** |

Our results show that the PCA+Decision Tree model is showing the least performance across all five performance metrics with recall value as lowest value indicating that the model misses some truly positive situations. When using RF, particularly when combined with PSO, there is an increased performance in most performance metrics compared to the case when Decision Tree was used. We observed, combination of both feature selection methods especially the PSO and transformer-based classifiers outperformed significantly the classification performance compared with traditional machine learning classifiers such as RF and Decision Tree. The PSO + Tab_transformer-based model has achieved the best accuracy (99.50%), precision (99.6%), recall (99.5%), F1-score (99.5%), and AUC (0.99). This model outperformed the other seven models in all performance metrics. The visualization of all evaluation results for eight different methods compared to each other are shown in Figure 4.

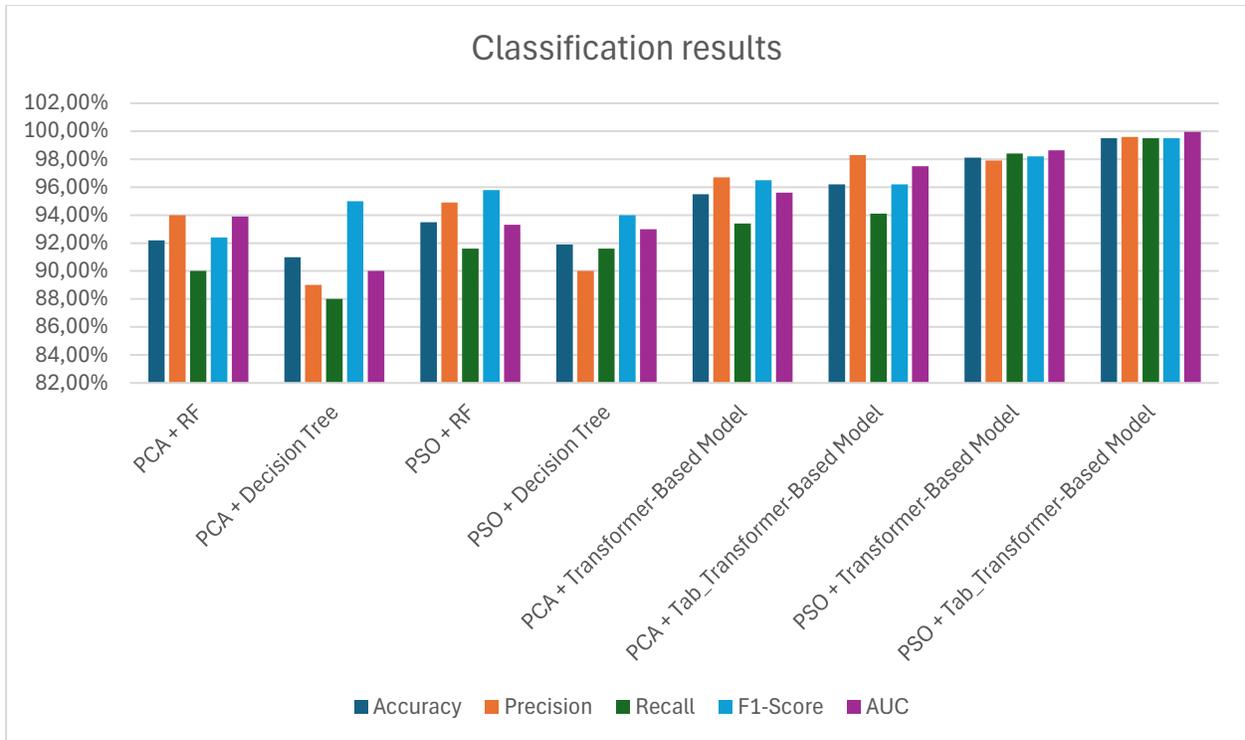

Figure 4: Visualization of the binary classification results in this paper for all eight experiments and using five performance metrics including accuracy, precision, recall, F1-score and AUC.

### 3.2. Analyzing details of the best performing model

As mentioned in Section 3.1, the best prediction results was achieved by PSO + Tab_transformer-based model. The outcome of PSO feature selection using this model is a reduced set of 45 features, selected based on their relevance to birth prediction and their ability to improve model performance. Each feature represents a critical aspect of the IVF dataset, categorized into groups such as Infertility Cause, Procedural Detail, Patient History, and Outcome. The selected featured and their explanation are described in Table 6.

Table 6. Explanation of each of the 45 features selected using PSO.

| Feature Name | Category | Description |
| --- | --- | --- |
| Cause of infertility - tubal disease | Infertility Cause | Indicates whether the patient's infertility is due to tubal disease. |
| Cause of infertility - partner sperm immunological factors | Infertility Cause | Refers to immunological issues with the partner's sperm that may affect fertility. |
| Cause of infertility - partner sperm morphology | Infertility Cause | Indicates abnormalities in sperm shape that may contribute to infertility. |
| Cause of infertility - endometriosis | Infertility Cause | Indicates whether the patient's infertility is caused by endometriosis. |
| Cause of infertility - female factors | Infertility Cause | Covers a range of female infertility factors not otherwise specified. |
| Cause of infertility - ovulatory disorder | Infertility Cause | Infertility is due to ovulatory disorders in the patient. |

| Feature Name | Category | Description |
| --- | --- | --- |
| Cause of infertility - patient unexplained | Infertility Cause | Referring to cases where the cause of infertility is unknown or unexplained. |
| Date of egg mixing | Procedural Detail | The date when eggs were mixed with sperm during the IVF procedure. |
| Date of embryo thawing | Procedural Detail | The date when frozen embryos were thawed for use in the IVF cycle. |
| Eggs mixed with donor sperm | Procedural Detail | Indicates whether donor sperm was used to mix with eggs during the IVF cycle. |
| Eggs mixed with partner sperm | Procedural Detail | It indicates whether the partner's sperm was used to mix with eggs during the IVF cycle. |
| Eggs thawed | Procedural Detail | Indicates the number of eggs that were thawed during the IVF procedure. |
| Embryos transferred | Procedural Detail | The number of embryos transferred to the uterus during the IVF cycle. |
| Embryos transferred from eggs micro-injected | Procedural Detail | Indicates embryos transferred that were developed from micro-injected eggs. |
| mbryos stored for use by patient | Procedural Detail | The number of embryos stored for future use by the patient. |
| Frozen cycle | Procedural Detail | Indicates whether the cycle involved frozen embryos. |
| Stimulation used | Procedural Detail | Indicates the type of hormonal stimulation protocol used during the cycle. |
| Total embryos created | Procedural Detail | The total number of embryos created during the IVF cycle. |
| Total number of previous treatments, both IVF and DI at clinic | Patient History | Total number of previous treatments (IVF or donor insemination) conducted at the same clinic. |
| Total number of previous IVF pregnancies | Patient History | The total number of pregnancies previously achieved through IVF. |
| Total number of previous DI pregnancies | Patient History | The total number of pregnancies previously achieved through donor insemination. |
| Feature Name | Category | Description |
| Total number of live births - conceived through IVF | Patient History | Total number of live births achieved through IVF treatment. |
| Total number of live births - conceived through IVF or DI | Patient History | The total number of live births achieved through either IVF or donor insemination. |
| Total number of previous pregnancies - IVF and DI | Patient History | Total number of previous pregnancies achieved through either IVF or donor insemination. |
| Donated embryo | Procedural Detail | Indicates whether a donated embryo was used during the IVF cycle. |
| Heart one delivery date | Outcome | The delivery date of the first baby (if multiple births occurred). |
| Heart one week's gestation | Outcome | The number of weeks of gestation for the first baby. |
| Heart two delivery date | Outcome | The delivery date of the second baby (if multiple births occurred). |
| Heart two weeks gestation | Outcome | The number of weeks of gestation for the second baby. |
| Total number of previous DI cycles | Patient History | Total number of donor insemination cycles performed prior to the current treatment. |
| Type of infertility - female secondary | Infertility Cause | Indicates secondary infertility in the female patient. |
| Type of infertility - male primary | Infertility Cause | Indicates primary infertility in the male partner. |
| PGD (Preimplantation Genetic Diagnosis) | Procedural Detail | Indicates whether PGD was performed to screen embryos for genetic abnormalities. |
| PGT-A treatment | Procedural Detail | Indicates whether Preimplantation Genetic Testing for Aneuploidy (PGT-A) was used. |

| PGT-M treatment | Procedural Detail | Indicates whether Preimplantation Genetic Testing for Monogenic disorders (PGT-M) was used. |
| --- | --- | --- |
| Total eggs mixed | Procedural Detail | The total number of eggs mixed with sperm during the IVF cycle. |
| Fresh eggs stored | Procedural Detail | Indicates whether fresh eggs were stored for future use. |
| Fresh eggs stored (0/1) | Procedural Detail | Binary indicator for whether fresh eggs were stored. |
| Total number of live births - conceived through IVF or DI | Outcome | Total number of live births achieved from either IVF or donor insemination cycles. |

The training and validation performance of the tab_transformer model when combined with PSO, our top-performing model in this study, is also shown in Figure 5.

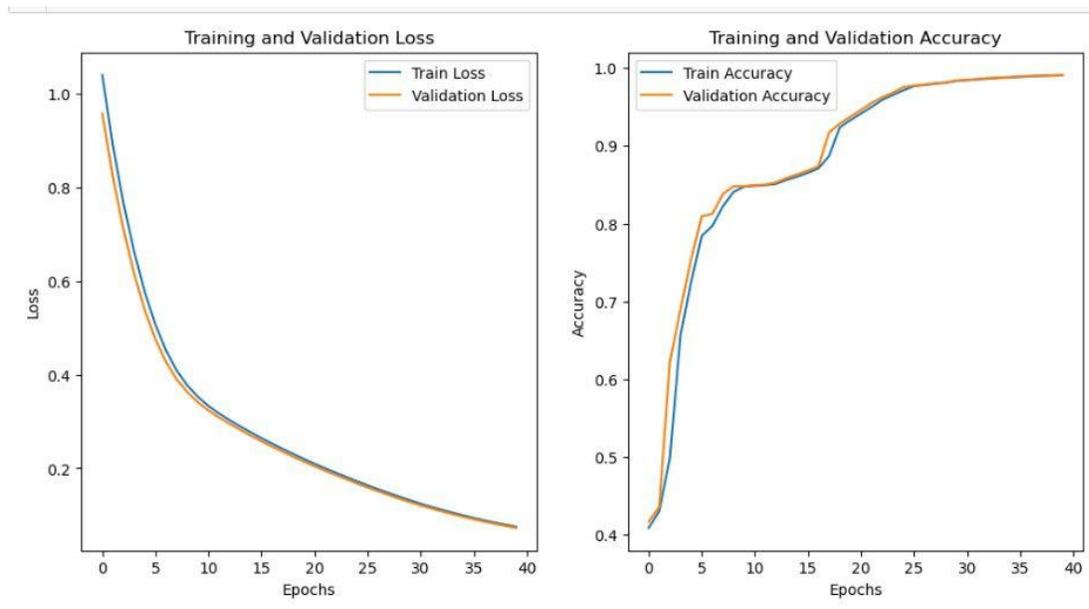

Figure 5: Training and Validation Performance of the tab transformer model Over 40 Epochs

Over 40 epochs, the tab_transformer model showed steady improvements in both training and validation metrics. Throughout the training process, both training and validation loss decreased gradually and stayed closely aligned, as seen in Figure 5's left panel. The close tracking between losses effectively minimizes overfitting while maintaining strong generalization to unknown data. The validation loss by the last epoch was 0.0729, which was very close to the training loss of 0.0753. The right panel displays the training and validation accuracy, demonstrating a steady improvement and convergence to more than 99% by the end of the epoch. These outcomes show how well the tab transformer generalizes and how robust its learning ability is.
As seen in Figure 5's right panel, during the first epochs, the model showed fast convergence, beginning at about 40% for both metrics. The validation accuracy reached a peak of 99.09% by epoch 40, which was almost the same as the training accuracy of 99.07%. The strong correlation between training and validation metrics highlights the robustness and dependability of the tab_transformer model.

## 4. Discussion and conclusion

In this study, we explored a variety of machine learning and deep learning models in combination of two feature selection techniques for predicting live birth success in IVF using the comprehensive HFEA dataset.

Our study could achieve a very high performance for five different evaluation metrics by utilizing PSO for feature selection combined with tab transformer, an advanced deep learning model. The AI pipeline is designed by integrating PSO for feature selection, the tab transformer for tabular data classification and attention mechanism, balanced datasets to address class imbalance, cross-validation to prevent overfitting, and robust regularization techniques to enhance model stability. The propose model then has the potential to deal with common problems like overfitting, inconsistent patient data, and uneven datasets thus showing promise for a clinically applicable tool for predicting live birth success in IVF.

With an accuracy of 99.5%, precision of 99.6%, recall of 99.5%, F1-score of 99.5%, and AUC of 0.99, the PSO + Tab_transformer-based model produced exceptional results, making it the most successful model for forecasting the success of live births. In contrast, the excellent accuracy and recall offered by the transformer-based techniques could not be achieved by models like PCA + Decision Tree and PCA + Random Forest, despite their effectiveness. These finding highlight that an enhanced IVF outcome prediction could be obtained by using deep learning-based transformer models. These deep learning models provide the advantage over conventional classifiers in terms of their ability to recognize relevant features and to capture intricate relationships within the data.

Particularly, the tab transformer offers several advantages over traditional models. First, it can efficiently handle the high-cardinality categorical features by learning embedding instead of one-hot encoding, which can lead to representations that are sparse and high-dimensional. Second, it captures complex interactions between features using self-attention, which traditional models might overlook. Third, it reduces the need for extensive manual feature engineering, enabling end-to-end learning directly from raw tabular data. This model is particularly effective in domains where meaningful relationships between features play a crucial role. When compared to traditional machine learning models, the tab transformer excels with larger datasets and high-cardinality features, offering state-of-the-art performance. Historically, the challenges in IVF outcome prediction also including live birth prediction included limited application of advanced deep learning models for tabular data. The application of these advanced deep learning techniques is terofre explored in this study for the first time. We also have surveyed some of previous efforts that used the HFEA dataset for the classification of live birth success as a binary outcome (success/failure), similar goal to our study (Table 7).

As shown in Table 7, the results of this research surpass all previous studies utilizing similar datasets. Notably, compared to the study by Sadegh-Zadeh et al. [21] achieving an accuracy of 96.35%, which employed the same dataset from 2010–2018 and adhered to same inclusion and exclusion criteria as used in our study, our study could improve upon these results with an accuracy of 99.50% and an AUC of 0.99%.

Future work could focus on integrating additional domain-specific features related to IVF treatments and patient characteristics to further improve model performance. Exploring the use of other advanced deep learning models, including those that account for sequential or temporal data, will be also explored for their potential to enhance prediction accuracy. Expanding the dataset to include more diverse populations and treatment types would help improve model generalizability and applicability in broader IVF contexts.

Table 7: Related works Using the HFEA Dataset for Predicting Live Birth Success

| Study | Dataset Used | Key Features/Methods | Model Used | Performance Metrics |
|---|---|---|---|---|
| Zhang et al. [15] | 57,558 NC-IVF cycles (2005–2016) | Patient demographics, hormonal profiles, cycle history, treatment outcomes; data balancing (SMOTE), SHAP, cross-validation | Artificial Neural Network (ANN) | F1-score: 70.87%, AUC: 0.7939 |
| Sadegh-Zadeh et al. [21] | 495,630 IVF cycles (2010–2018) | Clinical and demographic data; temporal validation, feature normalization, interpretability | Ensemble models (AdaBoost, LogitBoost) | Accuracy: 96.35% |
| McLernon et al.[16] | 113,873 women, 184,269 cycles (1999–2008) | Multiple complete IVF/ICSI cycles; pre- and post-treatment analysis | Discrete-time Logistic Regression | C-index: 0.73 (pre-treatment), 0.72 (post-treatment) |
| Jones et al. [17] | 93,495 women, 174,418 IVF cycles (1991–1998) | Focused on the likelihood of live birth success | Logistic Regression | AUC: 0.635 |
| Sanders et al. [18] | 190,010 IVF cycles (2016–2018) | Comparison of PGT-A and non-PGT-A cycles; odds ratios (ORs), descriptive statistics | Binary Logistic Regression | Focused on ORs instead of AUC |
| **This Research** | **2010–2018 IVF dataset** | **Advanced ML techniques, IVF-specific preprocessing** | **PSO + Tab Transformer** | **Accuracy: 99.50%, AUC: 99.96%** |

**Funding for this study:** This study was funded by ACMIT COMET Module FFG project (FFG number: 879733, application number: 39955962).

**Data availability statement**

The original contributions presented in the study are included in the article/supplementary material, further inquiries can be directed to the corresponding author.

**References**

[1] Morse, K. (2022). ni ve rs ity of e To w n ve rs ity e To w, 213.

[2] Liu, L., Liang, H., Yang, J., Shen, F., Chen, J., & Ao, L. (2024). Clinical data-based modeling of IVF live birth outcome and its application. *Reproductive biology and endocrinology*, *22*(1), 1–12. DOI:10.1186/s12958-024-01253-3

[3] Uyar, A., Sengul, Y., & Bener, A. (2015). Emerging technologies for improving embryo selection: a systematic review. *Advanced health care technologies*, 55. DOI:10.2147/ahct.s71272

[4] Hanassab, S., Abbara, A., Yeung, A. C., Voliotis, M., Tsaneva-Atanasova, K., Kelsey, T. W., … Dhillo, W. S. (2024). The prospect of artificial intelligence to personalize assisted reproductive technology. *Npj digital medicine*, *7*(1). DOI:10.1038/s41746-024-01006-x

[5] Patel, D. J., Chaudhari, K., Acharya, N., Shrivastava, D., & Muneeba, S. (2024). Artificial Intelligence in Obstetrics and Gynecology: Transforming Care and Outcomes. *Cureus*, *16*(7). DOI:10.7759/cureus.64725

[6] Vassakis, K., Petrakis, E., & Kopanakis, I. (2018). Big data analytics: Applications, prospects and challenges. *Lecture notes on data engineering and communications technologies*, *10*, 3–20. DOI:10.1007/978-3-319-67925-9_1

[7] Ueno, S., Berntsen, J., Ito, M., Okimura, T., & Kato, K. (2022). Correlation between an annotation-free embryo scoring system based on deep learning and live birth/neonatal outcomes after single vitrified-warmed blastocyst transfer: a single-centre, large-cohort retrospective study. *Journal of assisted reproduction and genetics*, *39*(9), 2089–2099. DOI:10.1007/s10815-022-02562-5

[8] Coticchio, G., Lagalla, C., Taggi, M., Cimadomo, D., & Rienzi, L. (2024). Embryo multinucleation: detection, possible origins, and implications for treatment. *Human reproduction*, *39*(11), 2392–2399. DOI:10.1093/humrep/deae186

[9] Bamford, T., Easter, C., Montgomery, S., Smith, R., Dhillon-Smith, R. K., Barrie, A., … Coomarasamy, A. (2023). A comparison of 12 machine learning models developed to predict ploidy, using a morphokinetic meta-dataset of 8147 embryos. *Human reproduction*, *38*(4), 569–581. DOI:10.1093/humrep/dead034

[10] Uyar, A., Bener, A., & Ciray, H. N. (2015). Predictive Modeling of Implantation Outcome in an in Vitro Fertilization Setting. *Medical decision making*, *35*(6), 714–725. DOI:10.1177/0272989X14535984

[11] Giscard d'Estaing, S., Labrune, E., Forcellini, M., Edel, C., Salle, B., Lornage, J., & Benchaib, M. (2021). A machine learning system with reinforcement capacity for predicting the fate of an ART embryo. *Systems biology in reproductive medicine*, *67*(1), 64–78. DOI:10.1080/19396368.2020.1822953

[12] Huang, B., Zheng, S., Ma, B., Yang, Y., Zhang, S., & Jin, L. (2022). Using deep learning to predict the outcome of live birth from more than 10,000 embryo data. *BMC pregnancy and childbirth*, *22*(1),

1–7. DOI:10.1186/s12884-021-04373-5

[13] Jiang, V. S., Kandula, H., Thirumalaraju, P., Kanakasabapathy, M. K., Cherouveim, P., Souter, I., … Shafiee, H. (2023). The use of voting ensembles to improve the accuracy of deep neural networks as a non-invasive method to predict embryo ploidy status. *Journal of assisted reproduction and genetics*, *40*(2), 301–308. DOI:10.1007/s10815-022-02707-6

[14] Kragh, M. F., & Karstoft, H. (2021). Embryo selection with artificial intelligence: how to evaluate and compare methods? *Journal of assisted reproduction and genetics*, *38*(7), 1675–1689. DOI:10.1007/s10815-021-02254-6

[15] Zhang, Y., Shen, L., Yin, X., & Chen, W. (2022). Live-Birth Prediction of Natural-Cycle In Vitro Fertilization Using 57,558 Linked Cycle Records: A Machine Learning Perspective. *Frontiers in endocrinology*, *13*(April), 1–12. DOI:10.3389/fendo.2022.838087

[16] McLernon, D. J., Steyerberg, E. W., Te Velde, E. R., Lee, A. J., & Bhattacharya, S. (2016). Predicting the chances of a live birth after one or more complete cycles of in vitro fertilisation: Population based study of linked cycle data from 113 873 women. *BMJ (online)*, *355*. DOI:10.1136/bmj.i5735

[17] Jones, C. A., Christensen, A. L., Salihu, H., Carpenter, W., Petrozzino, J., Abrams, E., … Keith, L. G. (2011). Prediction of individual probabilities of livebirth and multiple birth events following in vitro fertilization (IVF): A new outcomes counselling tool for IVF providers and patients using HFEA metrics. *Journal of experimental and clinical assisted reproduction*, *8*, 1–10.

[18] Sanders, K. D., Silvestri, G., Gordon, T., & Griffin, D. K. (2021). Analysis of IVF live birth outcomes with and without preimplantation genetic testing for aneuploidy (PGT-A): UK Human Fertilisation and Embryology Authority data collection 2016–2018. *Journal of assisted reproduction and genetics*, *38*(12), 3277–3285. DOI:10.1007/s10815-021-02349-0

[19] Hassan, M. R., Al-Insaif, S., Hossain, M. I., & Kamruzzaman, J. (2020). A machine learning approach for prediction of pregnancy outcome following IVF treatment. *Neural computing and applications*, *32*(7), 2283–2297. DOI:10.1007/s00521-018-3693-9

[20] Milewski, R. (2010). the Usage of Margin-Based Feature Selection Algorithm in Ivf Icsi / Et, *21*(34), 35–46.

[21] Sadegh-Zadeh, S. A., Khanjani, S., Javanmardi, S., Bayat, B., Naderi, Z., & Hajiyavand, A. M. (2024). Catalyzing IVF outcome prediction: exploring advanced machine learning paradigms for enhanced success rate prognostication. *Frontiers in artificial intelligence*, *7*(November), 1–18. DOI:10.3389/frai.2024.1392611

[22] Theng, D., & Bhoyar, K. K. (2024). *Feature selection techniques for machine learning: a survey of more than two decades of research*. , 66 Knowledge and Information Systems (Vol. 66). Springer London.

[23] Sowan, B., Eshtay, M., Dahal, K., Qattous, H., & Zhang, L. (2023). Hybrid PSO feature selection-based association classification approach for breast cancer detection. *Neural computing and applications*, *35*(7), 5291–5317. DOI:10.1007/s00521-022-07950-7

[24] Hammoumi, D., Al-Aizari, H. S., Alaraidh, I. A., Okla, M. K., Assal, M. E., Al-Aizari, A. R., … Bejjaji, Z. (2024). Seasonal Variations and Assessment of Surface Water Quality Using Water Quality Index (WQI) and Principal Component Analysis (PCA): A Case Study. *Sustainability (switzerland)* , *16*(13), 1–22. DOI:10.3390/su16135644

[25] Zhou, L., Xu, D., Yuan, Y., & Wang, L. (2024). Research on transformer fault intelligent diagnosis

technology based on improved random forest algorithm. *Journal of physics: conference series*, *2728*(1). DOI:10.1088/1742-6596/2728/1/012056

[26] Borji, A., Seifi, A., & Hejazi, T. H. (2023). An efficient method for detection of Alzheimer's disease using high-dimensional PET scan images. *Intelligent decision technologies*, *17*(3). DOI:10.3233/IDT-220315

[27] Charbuty, B., & Abdulazeez, A. (2021). Classification Based on Decision Tree Algorithm for Machine Learning. *Journal of applied science and technology trends*, *2*(01), 20–28. DOI:10.38094/jastt20165

[28] Tabinda Kokab, S., Asghar, S., & Naz, S. (2022). Transformer-based deep learning models for the sentiment analysis of social media data. *Array*, *14*(October 2021), 100157. DOI:10.1016/j.array.2022.100157

[29] Vaswani, A., Shazeer, N., Parmar, N., Uszkoreit, J., Jones, L., Gomez, A. N., … Polosukhin, I. (2017). Attention is all you need. *Advances in neural information processing systems*, *2017-Decem*(Nips), 5999–6009.

[30] Alzahrani, A. I. A., Al-Rasheed, A., Ksibi, A., Ayadi, M., Asiri, M. M., & Zakariah, M. (2022). Anomaly Detection in Fog Computing Architectures Using Custom Tab Transformer for Internet of Things. *Electronics (switzerland)*, *11*(23), 1–20. DOI:10.3390/electronics11234017

[31] Borji, A., Hejazi, T.-H., & Seifi, A. (2024). Introducing an ensemble method for the early detection of Alzheimer's disease through the analysis of PET scan images, 1–22.